# TARGETED ASPECT-BASED MULTIMODAL SENTIMENT ANALYSIS: AN ATTENTION CAPSULE EXTRACTION AND MULTI-HEAD FUSION NETWORK

*Jiaqian Wang\*, Donghong Gu\*, Chi Yang, Yun Xue¶, Zhengxin Song, Haoliang Zhao, Luwei Xiao*

School of Physics and Telecommunications Engineering
South China Normal University, Guangzhou 560001, China
{2018022015, gu_dh105}@m.scnu.edu.cn,
xueyun@scnu.edu.cn

**ABSTRACT**

Multimodal sentiment analysis has currently identified its significance in a variety of domains. For the purpose of sentiment analysis, different aspects of distinguishing modalities, which correspond to one target, are processed and analyzed. In this work, we propose the targeted aspect-based multimodal sentiment analysis (TABMSA) for the first time. Furthermore, an attention capsule extraction and multi-head fusion network (EF-Net) on the task of TABMSA is devised. The multi-head attention (MHA) based network and the ResNet-152 are employed to deal with texts and images, respectively. The integration of MHA and capsule network aims to capture the interaction among the multimodal inputs. In addition to the targeted aspect, the information from the context and the image is also incorporated for sentiment delivered. We evaluate the proposed model on two manually annotated datasets. the experimental results demonstrate the effectiveness of our proposed model for this new task.

## 1. INTRODUCTION

Sentiment analysis, also referred to as sentiment classification, aims to extract opinions from a large number of unstructured texts and classifying them into sentiment polarities, positive, neutral or negative [1]. To date, much of the work on sentiment analysis focuses on textual data [2]. Notably, with the advances of social media, it is significant to precisely capture the sentiment via the presence of different modalities (i.e. textual, acoustic and visual) [3-4]. Recent initiatives reveal that nearly 40% of reviews on cellphone in ZOL.com contain both text and image, which attract over 3 times the attention than the text-only reviews [2]. As such, the ability to analyze sentiment on multimodal data is most pronounced.

On current shopping and social platforms, seeing that the text and image information is taken to mutually reinforce and complement each other, models are dedicatedly devised to classify the sentiment polarity by using both kinds of data and their latent relation [5]. Recent publications report their achievements on the task of multimodal sentiment analysis. Xu et al. propose a Multi-Interactive Memory Network, together with a aspect-level multimodal sentiment analysis (ABMSA) dataset for model evaluation [2]. Yu et al. develop methods for target-oriented multimodal sentiment classification (TMSC) [5-6] by integrating the attention mechanisms and the pre-trained ResNet [7]. Experimental results show that an even higher accuracy can be obtained by incorporating the image into classical sentiment analysis.

On the other hand, sentiments towards different aspects of more than one entity are discussed in the same unit of text in many scenarios. targeted aspect-based sentiment analysis (TABSA) combines the challenges and the superiorities of aspect-based sentiment analysis and target-oriented sentiment analysis, and paves a way for greater depth of analysis. Namely, this task requires the detection of the aspect category and the sentiment polarity for a given targeted entity. According to Saeidi et al., TABSA caters for more generic text by making fewer assumptions with a more delicate understanding, which is both creative and practical for sentiment analysis [8].

In this work, we introduce a new task, namely Targeted aspect-based multimodal sentiment analysis (TABMSA), which indicates the integrating of multimodal information into TABSA to facilitate the sentiment analysis. That is, by exploiting information from texts and images, the sentiment classification result with a higher accuracy can be obtained. As illustrated in Table 1, there are three targets in the text: 'Dr Lucille Corti', 'Dr Lukwiya' and 'Uganda'. For targets 'Dr Lucille Corti' and 'Dr Lukwiya', the aspects contain 'event' and 'appearance'. Notably, the sentiment polarity for 'appearance' is positive according to the image while that for 'event' is negative according to the text. On this occasion, an approach to precisely capture the information of both texts and images is highlighted.

In this paper, we propose an Attention Capsule Extraction and Multi-head Fusion network (EF-Net) on the task of TABMSA. In our model, a bidirectional-GRU and multi-head self-attention mechanism is established for text semantic information encoding while the ResNet-152 model and capsule network is employed for dealing with the image, modality to sentiment delivering. Lastly, the multimodal representation, concatenating with the original semantic representation, is fed into the sentiment classifier. Experiments are conducted on two manually annotated multimodal datasets [5], which aims to verify the effectiveness of EF-Net comparing to the baseline method.

\* contribute equally
¶ corresponding author

**Table 1.** An example for TABMSA

| Image | Text | Target | Aspect |
|---|---|---|---|
| 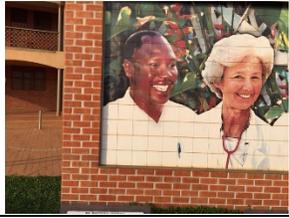 | What do health heroes look like? Dr Lucille Corti died AIDS 1996, Dr Lukwiya died Ebola 2000. 52yrs serving#Uganda | Dr Lucille Corti | [event] **negative** |
| | | | [appearance] **positive** |
| | | Dr Lukwiya | [event] **negative** |
| | | | [appearance] **positive** |
| | | Uganda | [place] **neutral** |

## 2. RELATED WORK

### 2.1. TABSA

Within TABSA, the sentiment associated to specific aspect of the entity is discussed. As mentioned above, current work is established on the foundation of Saeidi et al.'s studying of baseline method and the dataset. As an example, Ma et al. develop a LSTM-based model that utilizes the commonsense knowledge proposed in SenticNet for external knowledge incorporating [9]. Language processing models, such as BERT, is also taken as an alternative [10]. Besides, a recurrent entity network is designed and deployed to track entity state via word-level information and sentence-level hidden memory [11]. In some studies, researchers generally exploit the context-independence and randomly-initialized vectors to represent the aspects, which lack analyzing the interaction between aspect and its contexts.

### 2.2. ABMSA

The text-image pair is the most common form of multimodal data [2]. In most cases, the joint use of these modalities can not only enhance the sentiment expressing, but also improves the classification accuracy in sentiment analysis. As presented in [3], a co-memory attentional mechanism to interactively model the interaction between text and image is established, and thus to analyze the effects on one modality to the other. Motivated by the fine-grained sentiment analysis, Multi-Interactive Memory Network is proposed to learn the interactive influences between cross-modality data and the self-influences in single-modality data [2]. Likewise, models like TomBERT [5] and ESAFN [6] also receive great which can maintain more related information. For multimodal interaction and fusion, the multi-head attention network is applied to maximize the contribution of each attention due to their superiority in multimodal sentiment classification tasks.

## 3. METHODOLOGY

The task of TABMSA can be formulated as follows: given an image $I$ and a text sequence of n words $v^c = \{v_1^c, v_2^c, \cdots, v_n^c\}$, containing $m$ targets $v^t = \{v_1^t, v_2^t, \cdots, v_m^t\}$, to characterize the aspect $a$. Our purpose is to figure out the sentiment polarity $y$ towards ($v^t, a$) in ($v^c, I$).

The architecture of EF-Net model is shown in Fig.1. Our model mainly contains four layers: feature extracting layer, multimodal interaction layer, multimodality fusion layer and final classification layer. The model firstly extracts the features from texts and images and encodes them into corresponding representations in the feature extracting layer. Then multimodal information interaction is carried out to preserve the more-related information. In the multimodality fusion layer, the multi-head attention-based fusion network is performed to filter and fuse the inter-modal information. The multimodality fusion outcome, together with the original semantic sequences, is concatenated for final sentiment prediction. The details of each part are described as follows. We start with a brief introduction of the Multi-Head Attention (MHA) network, which is applied to our model.

### 3.1. Multi-Head Attention (MHA) network

The Multi-Head Attention (MHA) aims to perform multiple attention function in parallel, which can be considered as an improvement of the traditional attention mechanism [12]. Basically, a traditional attention is defined as:

$$Attention(Q, K, V) = \text{soft max}(\frac{QK^T}{\sqrt{d_k}}) \quad (1)$$

Where $Q$ stands for Query, $K$ for Key and $V$ for Value. The regulator $\sqrt{d_k}$ is taken to constrain the dot product value. In MHA, the inputs $Q$, $K$ and $V$ are mapped through the parameter matrices. Then the attention function is computed in parallel, whose outcomes are concatenated to obtain the multi-head attention value. Thus, we have

$$head_i = Attention(QW_i^Q, KW_i^K, VW_i^V) \quad (2)$$

$$MultiHead(Q,K,V) = Concatenate(head_1, \cdots head_n)$$

and where $W_i^Q$, $W_i^K$ and $W_i^V$ represent the projections parameter matrices for the corresponding inputs and $head_i$ is the attention of $i$-th head.

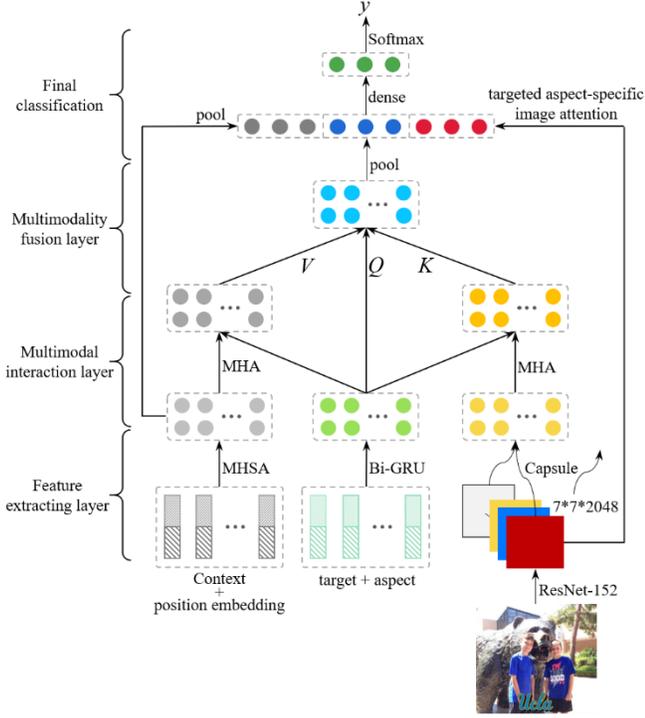

**Fig. 1.** Overall architecture of the proposed EF-Net.

In this work, we also employed Multi-head Self-Attention (MHSA), which can be regard as a special kind of MHA. In MHSA, the identical inputs are sent to the model, i.e. $Q = K = V$. In this way, the attention is delivered as:

$$MHSA = MultiHead(X, X, X) \quad (3)$$

where $X$ indicates a general input of the MHA network.

### 3.2. Feature extracting layer

In this layer, both the texts and the images are sent to the model as the inputs. For the textual data, we shall map the words into a low-dimensional vector by looking up in the pretrained Glove [12]. Thus, the word embeddings for the given text are obtained. Let $w^c = \{w_1^c, w_2^c, \cdots, w_n^c\}$ be the word embedding of the context, $w^t = \{w_1^t, w_2^t, \cdots, w_m^t\}$ be that of the target and $w^a$ be that of the aspect.

**Context representation.** In addition to the semantic information, the position information is also considered. The relative distances between each word and the target is computed and the outcomes are presented as the position embeddings $p^c = \{p_1^c, p_2^c, \cdots, p_n^c\}$. By employing the MHSA mechanism, the concatenation of word embeddings and the position embeddings is transferred to the context $h^c = \{h_1^c, h_2^c, \cdots, h_n^c\}$, which is delivered as:

$$h^c = MHSA([w^c, p^c])$$
$$= MultiHead([w^c, p^c],[w^c, p^c],[w^c, p^c]) \quad (4)$$

The context Representation retains the original semantic information and syntactic structure of the context to the greatest extent, so we get $h_{avg}^c$ after average pooling of $h^c$, which is used for feature fusion in the final sentiment classification:

$$h_{avg}^c = \sum_{i=1}^n h_i^c / n \quad (5)$$

**Targeted aspect representation.** Seeing that both target and aspect are short-sequence text, Bi-GRU is employed to capture the semantic information. That is, the word embeddings of target $w^t$ and aspect $w^a$ are concatenated and sent to Bi-GRU network. The targeted aspect representation $h^{ta} = \{h_1^{ta}, h_2^{ta}, \cdots, h_m^{ta}\}$ is given by:

$$\overrightarrow{h_j} = \overrightarrow{GRU}([w_j^t, w^a]), j \in [1, m] \quad (6)$$

$$\overleftarrow{h_j} = \overleftarrow{GRU}([w_j^t, w^a]), j \in [m, 1] \quad (7)$$

$$h^{ta} = [\overrightarrow{h_j}, \overleftarrow{h_j}], j \in [1, m] \quad (8)$$

**Visual representation.** In order to make full use of image information, a most effective image recognition method, ResNet-152, is used for image feature extraction. For a specific input image $I$, we re-size it to a 224×224-pixel image $I'$. With the pre-trained ResNet-152, the image feature vector $R$ can be:

$$R = \text{Res}Net(I') \quad (9)$$

where $R$ is a 7*7*2048 dimensional tensor.

Nevertheless, since ResNet is absent of tackling the position information of target in the image, we feed R into a one-layer capsule network. Thereby, the image representation $h^i$, which contains the position information of the target, is written as

$$h^i = Capsule(R) \quad (10)$$

**Targeted aspect specific image attention.** Aiming to remove the unrelated context to the target (e.g. image background) and preserve the most related part, the attention mechanism is applied, based on which the more essential image representation $h_{att}^i$ is defined as:

$$h_{att}^i = Attention(h_{avg}^{ta}W^{ta}, RW^R, RW^R) \quad (11)$$

where $W^{ta}$ and $W^R$ are trainable parameter matrices for mapping $h_{avg}^{ta}$ and $R$ into sub-space of the same dimension.

## 3.3. Multimodal interaction layer

The multimodal interaction layer is responsible for analyzing the relation between the targeted aspect, context and the image, respectively, and thus distilling the key information from the multimodal inputs. The main purpose of this layer is to obtain the targeted aspect specific textual attention and the targeted aspect specific visual attention. Therefore, MHA network is utilized to understand the interaction with respect to the targeted aspect.

For the targeted aspect $h^{ta}$ and the context $h^c$, we set $h^{ta}$ as $Q$ and $h^c$ as $K$. With $K = V$, the interaction of between the targeted aspect and its context is characterized by:

$$h^{tac} = MultiHead(Q, K, V) = MultiHead(h^{ta}, h^c, h^c) \quad (12)$$

where $h^{tac}$ is the targeted aspect specific context representation.

Likewise, the representation of targeted aspect specific image $h^{tai}$, indicating the interaction of between the targeted aspect and the image is

$$h^{tai} = MultiHead(Q, K, V) = MultiHead(h^{ta}, h^i, h^i) \quad (13)$$

## 3.4. Multimodality fusion layer

In practical use, not only the targeted aspect, but also the context and the image contain the sentiment information for determining sentiment polarities. Accordingly, following the multimodal interaction layer, the targeted aspect-specific representations from different modalities are incorporated. Instead of using a gated mechanism to control the contribution of each component, we tend to take the three representations as the inputs of MHA model for information fusion. By exploiting the MHA mechanism, the multimodal representation is then given as:

$$h^{taci} = MultiHead(Q, K, V) = MultiHead(h^{ta}, h^{tac}, h^{tai}) \quad (14)$$

where $h^{taci}$ stands for the multimodal representation. In eqn.(14), we set $h^{ta}$ as $Q$, $h^{tac}$ as $K$ and $h^{tai}$ as $V$.

Based on the multimodal representation $h^{taci}$, we can calculate $h^{taci}_{avg}$ via the average pooling (eqn.(15)). This obtained representation is further enriched by concatenating the average representation of the context and the image ($h^c_{avg}$ and $h^i_{att}$).

$$h^{taci}_{avg} = \sum_{i=1}^{n} h^{taci}_i / n \quad (15)$$

$$O = [h^c_{avg}, h^{taci}_{avg}, h^i_{att}] \quad (16)$$

where $O$ is the final representation with multimodal information.

## 3.5. Final Classification

The aforementioned final representation $O$ is fed into softmax classifier for sentiment polarity distribution identification, which is

$$x = W_O^T O + b_O \quad (17)$$

$$\tilde{y} = \text{soft max}(x) = \frac{\exp(x)}{\sum_{k=1}^{c} \exp(x)} \quad (18)$$

where $W_O^T$ and $b_O$ is the trainable weight matrix and offset vector, and $C$ is the number of sentiment polarities.

## 3.6. Model training

The training process is conducted on by using the categorical cross-entropy, which is expressed as:

$$L = -\sum_i^m \sum_j^C y_i^j \log \tilde{y}_i^j + \lambda \|\theta\|^2 \quad (19)$$

where $m$ is the number of aspect terms in the sentence, C is the number of sentiment polarities. The parameter $y_i$ stands for the real sentiment distribution of $i$-th aspect term and $\tilde{y}_i^j$ is the predicted one on $j$-th sentiment polarity. Besides, $\lambda$ is the weight of $L_2$ regularization.

## 4. EXPERIMENT

### 4.1. Dataset

We manually annotate a large-scale TABMSA dataset based on two publicly available TMSC datasets, Twitter15 and Twitter17 [5]. Three experienced researchers, who work on natural language processing (NLP), are invited to extract targets and aspects in the sentences and label their sentiment polarities. To start with, 500 samples from dataset are randomly picked in advance to reveal the most appearing target and aspect types, which are 'people', 'place', 'time', 'organization' and 'other'. The targets, as well as the corresponding aspects, are presented in Table 2. In such manner, the annotated Twitter15 contains 3259 samples for training, 1148 for validation and 1059 for testing while the corresponding numbers in Twitter 17 are 3856, 1240 and 1331.

**Table 2.** targets and their respective aspects

| Target | Aspect | Target | Aspect |
|---|---|---|---|
| people | general event phenomenon environment experience other | place | general appearance achievement event speech other |
| organization | general event other | time | general event other |

**Table 3.** Statistics of Twitter15 and Twitter17 dataset

| Attribute | Twitter15 | Twitter17 |
|---|---|---|
| #Sentence | 3502 | 2910 |
| #Label | 3 | 3 |
| #Target aspect pair | 5466 | 6427 |
| Avg. of #Aspect/Sentence | 1.6 | 2.2 |
| Avg. text length/ Sentence | 13.2 | 13.9 |
| Max text length/ Sentence | 36 | 31 |
| Min text length/ Sentence | 1 | 3 |

Considering the TABMSA task, each sample from our dataset is composed of images and texts, together with targets and aspects of specific sentiment polarities. The expressed sentiment polarities are predefined as positive, neutral or negative. Details of our dataset is exhibited in Table 3.

### 4.2. Experimental setting

As mentioned above, experiments are conducted on dedicatedly-annotated datasets for working performance evaluation. We set the maximum padding length of textual content as 36 for Twitter15 and 31 for Twitter17. The images are sent to pre-trained ResNet-152 to obtain the 7*7*2048-dimension visual feature vector. For our model, we set the learning rate as 0.001, the dropout rate as 0.3 and the batch size as 128. The attention head number is 4.

### 4.3. Model comparison

In order to verify the superiority of our model, we separately compare our model with classical textual sentiment analysis methods (LSTM, GRU, ATAE-LSTM, MemNet and IAN) and the representative multimodal sentiment analysis methods (Res-MemNet and Res-IAN).

**LSTM** is taken to detect the hidden states of the context. As a lighter version of LSTM, **GRU** has simple model structure and strong capability of modeling long-term sequences of texts. **ATAE-LSTM** [1] applies LSTM and concatenating process to get the aspect embeddings while the attention network aims to select the word of sentiment significance. **MemNet** [13] applies a multi-layer attention mechanism on top of the common word embedding layer. The representations in **IAN** [14] are modeled on the foundation of the LSTM based interactive attention networks. And hidden states are taken to compute the attention scores by the pooling process. **Res-MemNet and Res-IAN** take the max-pooling layer of ResNet and the hidden representation of MemNet or IAN to concatenate for multi-modality sentiment classification. Notably, for all the aforementioned model, the sentiment polarity distribution of the target is finally determined by using the Softmax classifier.

### 4.4. Main Results

In this experiment, we adopt accuracy and Macro-F1 as evaluation metrics to denote the working performance. Table 4 shows the main results. In the classical TABSA tasks, the proposed model, which removes the image processing part, labeled as 'EF-Net (Text)', has the best and most consistent outcomes on two datasets. Among all the models, LSTM obtains the worst performance due to its lack of distinguishing targets and contexts in the sentence. In comparison, with the analysis of target and aspect, the working performance is considerably optimized. Besides, the employment of attention mechanism also contributes to the classification accuracy improvement. The EF-Net (Text) make use of both the position information and the semantic information. In this way, the representations in our model are considerably more informative for delivering sentiment. Furthermore, the MHA network captures the interaction between targeted aspect and context, based on which more essential information is preserved for sentiment classification.

On the other hand, the multimodal sentiment analysis models are generally more competitive than the basic textual sentiment analysis ones. With the integration of visual context, an even higher classification accuracy is thus accessible. On the task of TABMSA, EF-Net still significantly outperforms the baseline models. The minimum performance gap of 1.89% and 0.9% for Twitter15 and Twitter17 can be observed in Table 4 against the Res-EF-Net (Text) method. Clearly, our model is a better alternative for the task of multimodality sentiment analysis. In spite the effectiveness of EF-Net (Text), another explanation is that we fuse the image data into the texts, together with investigating the multimodal interaction, which exploits the sentiment information and the relation of multimodalities. Since EF-Net is more capable of dealing with the TABMSA tasks, it is reasonable to expect even higher accuracy in more evaluation settings, as it is the case.

**Table 4.** Comparative Results of EF-Net and Baselines

| | Method | Twitter15 | | Twitter17 | |
|---|---|---|---|---|---|
| | | ACC | F1 | ACC | F1 |
| Text | LSTM | 62.61 | 49.38 | 60.55 | 55.00 |
| | GRU | 67.99 | 58.43 | 64.83 | 62.41 |
| | ATAE | 68.17 | 59.09 | 65.28 | 62.89 |
| | MemNet | 69.49 | 63.10 | 65.96 | 62.80 |
| | IAN | 70.44 | 63.64 | 63.64 | 59.74 |
| | EF-Net(Text) | **71.67** | **67.30** | **66.19** | **62.21** |
| Text + Visual | Res-MemNet | 64.30 | 56.64 | 62.58 | 59.32 |
| | Res-IAN | 64.11 | 57.87 | 60.70 | 55.91 |
| | Res-EFNet(Text) | 71.76 | 62.13 | 66.87 | 63.51 |
| | EF-Net | **73.65** | **67.90** | **67.77** | **65.32** |

### 4.5. MHA head number impact analysis

We further studied the head number of MHA for better obtain the relation among modalities. At this stage, we vary the number of head within the collection {1, 2, 3, 4, 5, 6}. The results on Twitter 15 and Twitter 17 of different head numbers are shown in Fig. 2. It can see that our model has the highest accuracy on head number of 4. For a smaller head number (i.e. 1,2,3), the MHA fails to maintain the significant

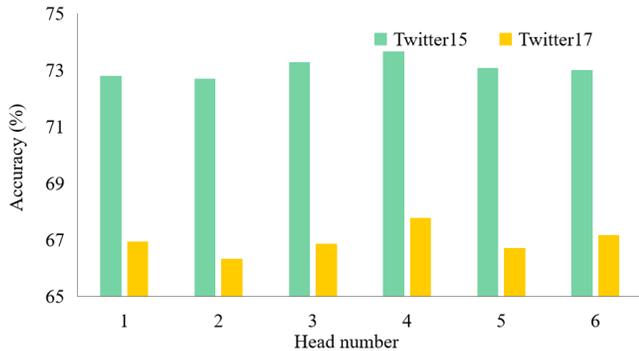
**Fig. 2.** Results of head number

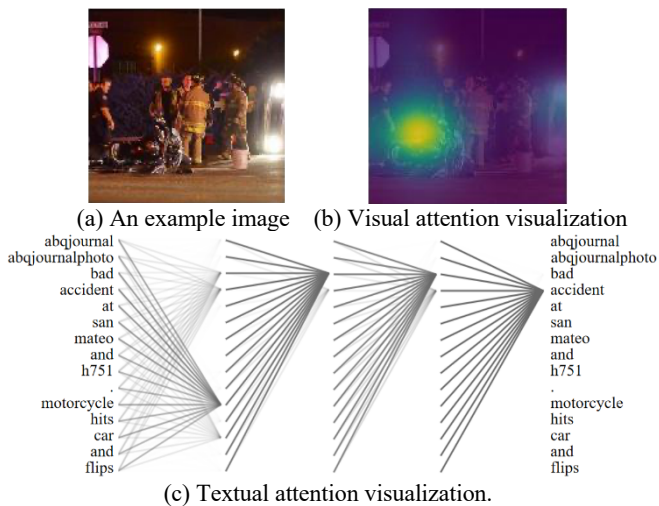
(a) An example image    (b) Visual attention visualization
(c) Textual attention visualization.
**Fig. 3.** An example of visual and textual attention

information, especially for long texts. Due to the parameter increasing and the model overfitting problem, the classification accuracy drops with the head number continues increasing (i.e. 5,6).

## 5. CASE STUDY

Fig.3 shows an example of visual and textual attention visualization. For the text '@ABQJournal Bad accident at San Mateo and H751. Motorcycle hits car and flip', the corresponding image is presented as Fig.3.(a). The target and aspect in the sentence are 'San Mateo' and 'event', respectively. According to Fig.3.(b), we can see that our model pays more attention to the motorcycle within the image. In addition, the MHA model (with Head=4), assigns more attention weights to words like 'Motorcycle', 'bad' and 'accident', as shown in Fig.3.(c). At this stage, our model classify the sentiment polarity as negative, which demonstrates that our model can properly capture the information and interaction of multimodalities.

## 6. CONCLUSION

In this work, we present a novel multimodal sentiment analysis task, namely TABMSA. As such, in line with the TABMSA tasks, the EF-Net model is designed and deployed. We first construct the representations of multimodal inputs. By employing the MHA network, the interaction between different representations is precisely captured to deliver more-related information. Moreover, the targeted aspect representation is enriched with the fusion of context and image information, which improves the multimodal sentiment classification accuracy to a large extent. Experiments results validate that the proposed model stably outperforms the baseline models.

## 7. REFERENCES

[1] Wang, Y., Huang, M., Zhu, X., & Zhao, L. (2016, November). Attention-based LSTM for aspect-level sentiment classification. In Proceedings of the 2016 conference on empirical methods in natural language processing (pp. 606-615).

[2] Nan Xu, Wenji Mao, Guandan Chen, Multi-Interactive Memory Network for Aspect Based Multimodal Sentiment Analysis, The Thirty-Third AAAI Conference on Artificial Intelligence (AAAI-19), 2019: 371-378.

[3] Yu, Y.; Lin, H.; Meng, J.; and Zhao, Z. 2016. Visual and textual sentiment analysis of a microblog using deep convolutional neural networks. Algorithms 9(2):41.

[4] Xu, N.; Mao,W.; and Chen, G. 2018. A co-memory network for multimodal sentiment analysis. In Proceedings of the 41st International ACM SIGIR Conference on Research and Development in Information Retrieval, 929–932.

[5] Jianfei Yu, Jing Jiang, Adapting BERT for Target-Oriented Multimodal Sentiment Classification, Proceedings of the Twenty-Eighth International Joint Conference on Artificial Intelligence, 2019: 5408-5414.

[6] Jianfei Yu, Jing Jiang, Rui Xia, Entity-Sensitive Attention and Fusion Network for Entity-Level Multimodal Sentiment Classification, IEEE/ACM Transactions on Audio, Speech, and Language Processing, 2020, Vol. 28: 429-439.

[7] He, K., Zhang, X., Ren, S., & Sun, J. (2016). Deep residual learning for image recognition. In Proceedings of the IEEE conference on computer vision and pattern recognition (pp. 770-778).

[8] Saeidi, M., Bouchard, G., Liakata, M., & Riedel, S. (2016). Sentihood: Targeted aspect based sentiment analysis dataset for urban neighbourhoods. arXiv preprint arXiv:1610.03771.

[9] Yukun Ma, Haiyun Peng, and Erik Cambria. 2018. Targeted aspect-based sentiment analysis via embedding commonsense knowledge into an attentive LSTM. Thirty-Second AAAI Conference on Artificial Intelligence.

[10] Sun C, Huang L, Qiu X. Utilizing BERT for aspect-based sentiment analysis via constructing auxiliary sentence[J]. arXiv preprint arXiv:1903.09588, 2019.

[11] Zhihao Ye, Zhiyong Li, A Variant of Recurrent Entity Networks for Targeted Aspect-Based Sentiment Analysis, 24th European Conference on Artificial Intelligence - ECAI 2020.

[12] Jeffrey Pennington, R., & Manning, C. GloVe: Global Vectors for Word Representation.
[13] Tang, D., Qin, B., & Liu, T. (2016, November). Aspect Level Sentiment Classification with Deep Memory Network. In Proceedings of the 2016 Conference on Empirical Methods in Natural Language Processing (pp. 214-224).
[14] Ma, D., Li, S., Zhang, X., & Wang, H. (2017, August). Interactive attention networks for aspect-level sentiment classification. In Proceedings of the 26th International Joint Conference on Artificial Intelligence (pp. 4068-4074).